\documentclass[letterpaper, 10 pt, journal, twoside]{IEEEtran}  % Comment this line out if you need a4paper
\usepackage{amsmath}
\usepackage{amssymb}
\usepackage{array}
\usepackage[dvipsnames]{xcolor}
\usepackage{graphicx}
\usepackage{epstopdf} 
\usepackage[caption=false,font=footnotesize]{subfig}

\usepackage{textcomp}
\usepackage{verbatim}
\usepackage{float}
\usepackage{booktabs}
\usepackage{multirow}

\usepackage{cite}
\usepackage{url} % 建议启用
\usepackage{xurl}

\usepackage{algorithm}
\usepackage{algorithmic} % 建议改用 algorithmicx

\makeatletter
\let\NAT@parse\undefined % 解决可能的冲突
\makeatother
\usepackage[colorlinks=true,
            linkcolor=green,
            citecolor=blue,
            urlcolor=black]{hyperref} % 添加 urlcolor

\IEEEoverridecommandlockouts                         

\title{\LARGE \bf{
Learning Robot Visual Navigation in Crowds via Intention-Aware \\ Scene Representations}
}

\usepackage{orcidlink}
\author{
    Han~Bao$^{\dag}$, 
    Bingyi~Xia$^{\dag}$,  
    Hanjing~Ye, 
    Yu~Zhan,
    Hao~Cheng,
    Baozhi~Jia,
    Wenjun~Xu,
    Jiankun~Wang
\thanks{Corresponding authors:  Baozhi Jia, Wenjun Xu, Jiankun Wang.}
\thanks{$^\dag$ Equal contribution.}
\thanks{Han Bao, Bingyi Xia, Hanjing Ye, Yu Zhan, Hao Cheng and Jiankun Wang are with the Shenzhen Key Laboratory of Robotics Perception and Intelligence, Department of Electronic and Electrical Engineering, SUSTech, Shenzhen, China (e-mail: \url{wangjk@sustech.edu.cn}).}
\thanks{Jiankun Wang is also with the Jiaxing Research Institute, SUSTech, Jiaxing, China.}
\thanks{Baozhi Jia is with Xiamen Key Laboratory of Visual Perception Technology and Application, and the Algorithm Research Center at Reconova Information Technology Co., Ltd. in Xiamen, China(e-mail: jiabaozhi@reconova.com).}    
\thanks{Wenjun Xu is with the Research Institute of MA\&EI, Peng Cheng Laboratory, Shenzhen, China(e-mail: xuwenjunwendy@gmail.com).}  
\thanks{Our code and appendix are available at https://broln7.github.io/socialbev.io/.}
}

\makeatletter
\def\ps@firstpage{%
  \ps@headings
  \def\@oddhead{\hbox{}\@IEEEheaderstyle\leftmark\hfil\thepage}%
}
\makeatother
\begin{document}
\markboth{IEEE ROBOTICS AND AUTOMATION LETTERS. PREPRINT VERSION. MARCH, 2026}
{Bao \MakeLowercase{\textit{et al.}}: Learning Robot Visual Navigation in Crowds via Intention-Aware Scene Representations}
\newcommand{\rv}[1]{\textcolor{black}{#1}}
\maketitle
\thispagestyle{firstpage}
\pagestyle{headings}

%%%%%%%%%%%%%%%%%%%%%%%%%%%%%%%%%%%%%%%%%%%%%%%%%%%%%%%%%%%%%%%%%%%%%%%%%%%%%%%%
\begin{abstract}
Robot crowd navigation requires the ability to infer human intentions while accounting for the structural constraints of the environment.
Currently, deep reinforcement learning (DRL) provides a promising method for learning navigation policies that understand human intentions.
However, most of them rely on limited scene representations, treating pedestrians as simple 2D points and ignoring rich visual cues from both humans and the environment.
To address this issue, iCrowdNav, a novel visual crowd navigation method with intention-aware scene representations, is introduced to encode behavioral and structural context from egocentric visual observations.
Our method employs two key components: a spatio-temporal encoder for extracting occupancy features of the scene, and Intent-Interact Former (I$^2$Former), an attention-based module that encodes human poses to infer pedestrians’ motion intentions.
These features are integrated into a compact state embedding that supports effective DRL policy training. 
Extensive experiments show that our method achieves superior performance over baselines, and real-world deployment demonstrates vision-based crowd navigation. 
% More details are available at \url{https://icrowdnav.github.io/}.

\end{abstract}

\begin{IEEEkeywords}
Human-aware motion planning, vision-based navigation, collision avoidance.
\end{IEEEkeywords}

%%%%%%%%%%%%%%%%%%%%%%%%%%%%%%%%%%%%%%%%%%%%%%%%%%%%%%%%%%%%%%%%%%%%%%%%%%%%%%%%
% ==========MAIN==========
\section{Introduction}

\begin{figure}[!t]
\centering
    \includegraphics[width=1.0\linewidth]{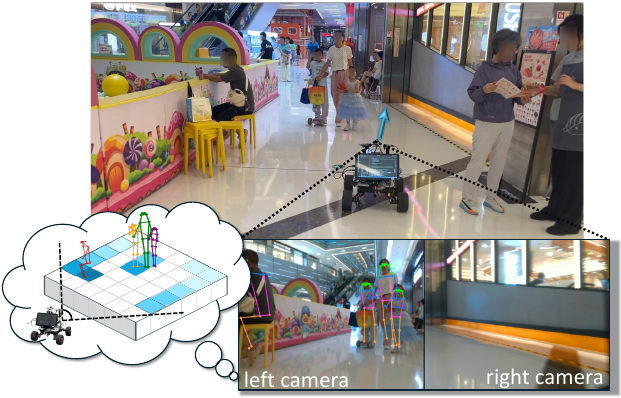}
    \caption{
    The robot is navigating in a shopping mall, which is required to avoid pedestrians in constrained space. 
    Our approach allows the robot to extract visual cues from onboard cameras for safe and efficient navigation.}
    \label{fig:cover}
    \vspace{-0.5cm}
\end{figure}

\IEEEPARstart{A}{utonomous} navigation has been significantly advanced by developments in visual perception and intelligent planning, paving the way for safe and reliable navigation in human presence.
However, relying solely on egocentric vision to navigate in dynamic, unstructured environments with dense crowds remains an open challenge \cite{lu2025fapp, ye2025RPF, sun2025namr, pengjianwei}. 
For instance, a service robot navigates through an unfamiliar shopping mall to perform delivery tasks, as shown in Fig.~\ref{fig:cover}.
Although this task is trivial for humans, robots still face severe difficulties in maneuvering through the dynamic environments safely and efficiently.
This requires understanding human motion intentions from visual input while accounting for environmental constraints, which enables appropriate and foresighted decision-making in dynamic crowds. 
Therefore, it is crucial to investigate vision-based crowd navigation methods that can operate reliably in everyday dense crowd environments.
% This requires a comprehensive understanding of social contexts from visual input to make appropriate and foresighted decisions that adhere to social norms.
% It is therefore crucial to investigate vision-based socially compliant navigation methods to promote their practical application in everyday scenarios.

Recently, deep reinforcement learning (DRL) approaches\cite{chen2019crowd,everett2018motion,liu2023intention} have shown promising performance in learning 
crowd dynamics and interactions
that are difficult to model explicitly. 
Nevertheless, the effectiveness of such approaches critically depends on the design of the state embeddings, which should preserve environmental cues that are both detailed and precise enough for optimal decision making.
Existing approaches \cite{liu2023intention, samsani2021socially, liu2020robot, xie2023drl} typically oversimplify the scene representation: for pedestrians, they use low-level states such as position and velocity on the 2D plane; for the environment, they rely on 2D binary occupancy maps or 2D single-line lidar scans.
These methods ignore rich visual cues in the raw images---such as the fact that humans may turn their heads and shoulders before changing their walking directions---and instead assume that intentions only concern the trajectories of pedestrians.
We argue that such low-level scene representations primarily overlook two key aspects: \textbf{(1) subtle yet critical human behaviors that express their intentions} (e.g., gestures, gazes, body poses), and \textbf{(2) spatial and semantic features of the environment}.
These factors are particularly crucial in the real world, such as the constrained passages shown in Fig.~\ref{fig:cover}, where a robot must rapidly infer the human intention to proactively yield for human crossing, and avoid collisions with walls or furniture.
As a result, low-level scene representations are limited in laboratory settings with certain participants, creating a gap to real-world scenarios.

In this article, we investigate the following question: \textit{How to learn scene representations from egocentric vision that better preserve visual cues for crowd navigation policy training? }
A direct end-to-end mapping from raw images to robot actions suffers from the curse of dimensionality and provides no guarantee that social contexts are effectively identified\cite{dugas2022navdreams}. 
Instead, Bird’s-Eye View (BEV) feature provides a compact yet informative scene representation by unifying visual inputs across multiple views and frames, which is convenient for downstream planning\cite{ma2024vision}.
It preserves spatial and semantic features of both the environment and pedestrians at an absolute scale while reducing dimensionality.
However, in highly dynamic and crowded scenarios, representing occupancy alone is insufficient, since human behaviors are often highly reactive and do not follow deterministic rules. 
Thus, we argue that human intention reasoning should be incorporated to enable foresighted decision-making.
Such intentions refer to pedestrians’ motion tendencies, for example whether they tend to walk straight or suddenly change direction. 3D human poses reflect these motion intentions and provide clear and reliable visual cues for inferring them\cite{salzmann2023robots,gao2025social}. We aim to capture such visual cues and incorporate them into our navigation framework.
To this end, we propose \textbf{iCrowdNav}, a visual navigation method that incorporates intention-aware scene representations by augmenting standard BEV with human pose features to facilitate the learning of crowd navigation policy.
Specifically, we design a spatio-temporal encoder that extracts occupancy features from a sequence of RGB-D observations of the dynamic environment. 
Moreover, we introduce the Intent-Interact Former (\textbf{$\text{I}^2$Former}), an attention-based module that learns implicit joint-level features from 3D human poses,  enabling the robot to infer human intentions.
By concatenating the BEV occupancy features with the intention-aware features extracted by $\text{I}^2$Former, we construct the intention-aware scene representations, which are then fed into the DRL framework to enable vision-based crowd navigation.
The main contributions of this article are as follows:
\begin{itemize}
    \item A novel visual encoder is incorporated in a DRL policy for crowd navigation using RGB-D cameras. It is end-to-end trained in simulation and achieves zero-shot sim-to-real deployment.

    \item The proposed intention-aware scene representations can implicitly capture behavioral and environmental contexts.
    Our method encodes BEV representations to densely build occupancy features of the scene, and leverages the attention mechanism to infer the human intention from 3D pose, prioritizing the human-robot interactions.

    \item We develop diverse human-centric environments in Isaac Sim, providing rich visual signals for training and benchmarking navigation in complex and dynamic scenarios. 
    Comprehensive experiments show that our approach outperforms existing state-of-the-art methods in crowd navigation performance.
\end{itemize}

\section{Related Work}

Robot navigation in dynamic crowds aims for robots to respond proactively and appropriately to surrounding pedestrians during navigation, achieving not only safety but also comfort and legibility\cite{survey2024,song2024vlm,payandeh2024social,luo2025gson}. 
Compared with rule-based approaches\cite{helbing1995social, fox2002dynamic}, DRL offers the advantage of optimizing long-term rewards that balance efficiency and safety while implicitly accounting for pedestrian comfort. 
For instance, DRL approaches\cite{everett2018motion, chen2019crowd} can overcome the frozen robot problem in dense crowds. 
Furthermore, coupled prediction and planning frameworks are developed to minimize discomfort to pedestrians by predicting their behavior. 
For example, human-robot interactions rules can be implicitly learned through graph neural network (GNN)\cite{liu2023intention} or integrated with the reward design\cite{samsani2021socially}.
However, these policies typically assume fully observable environments and often simplify the scene, thus showing limited generalization to various scenarios.

Real-world social navigation requires a more comprehensive understanding of holistic social contexts, which motivates researchers to explore state representations for integrating additional social cues\cite{zhou2022human}.
On one hand, the motion patterns of humans and robots are both constrained by the environment.
For example, DRL policies can be augmented with a map encoder that leverages pre-built grids maps for collision checking\cite{liu2020robot, yao2021crowd}.
Furthermore, real-time laser scans are used for the precise localization and prediction of dynamic obstacles\cite{xie2023drl, cui2021learning}.
On the other hand, simplistic safety distance approximations lead to overly conservative policies focused solely on obstacle avoidance.
Hence, researchers have exploited human behavioral knowledge to capture spatial relationships between humans and robots more faithfully.
% Proxemics based methods, construct risk maps from probabilistic reachable sets\cite{yang2023rmrl}, model the occupancy of pedestrians and obstacles with bounding capsules\cite{Zhu2023Collision}.
% Yang et al. construct risk maps from probabilistic reachable sets and modify the value network\cite{yang2023rmrl}.
% Zhu et al. model the occupancy of pedestrians and obstacles with bounding capsules\cite{Zhu2023Collision}.
Proxemics-based methods employ diverse representations such as probabilistic reachable sets \cite{yang2023rmrl} or bounding capsules for modeling pedestrian occupancy \cite{zhu2023collision}.
Other approaches consider physical constraints on human motion. 
For example, gait variations detected on 2D laser scans have been exploited as the state embeddings\cite{guldenring2020learning}.

Despite their progress, the above approaches primarily rely on low-dimensional planar observations that represent pedestrians to points, thereby discarding crucial visual cues. 
In contrast, this article leverages 3D vision features to benefit situation awareness in crowd navigation.
Although onboard cameras provide sufficient information about the surroundings, such visual inputs introduce challenges of partial observability and the curse of dimensionality in RL training. 
BEV representations have been widely adopted in autonomous driving and robotic navigation~\cite{jiang2024bevnav, hu2021fiery} due to their ability to unify multi-view observations and maintain spatio-temporal consistency.
However, in human-populated environments, current BEV representations are insufficient to capture visual signals that correlate with human behavior.
Previous trajectory prediction studies have revealed that human motion priors, such as gait patterns or human poses, can reliably reflect their intentions\cite{salzmann2023robots, gao2025social}. 
Inspired by these insights, we design the intention-aware scene representations, which integrate BEV representations with visual features capturing human intentions. These representations enable the DRL policy to more proactively and reliably avoid collisions in dense crowds.
\begin{figure*}[!t]
\vspace{0.25cm}
\centering
    \includegraphics[width=0.95\linewidth]{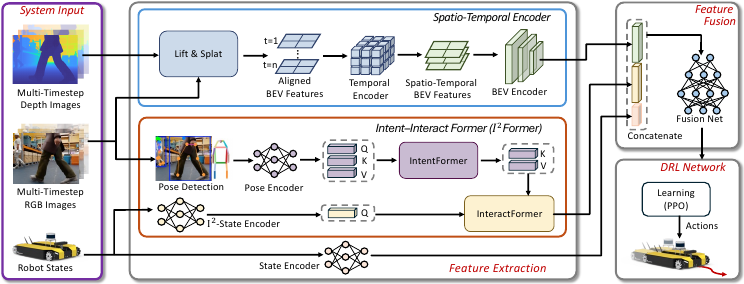}
    \caption{
    Our method consists of three primary components: a feature extraction module, a feature fusion module, and a DRL network. 
    It takes multi-timestep RGB-D images, pedestrian poses, and the robot’s internal states as inputs. 
    In the representation encoding stage, the spatio-temporal encoder and the I$^2$Former extract intention-aware scene representations, which are then fused with the robot’s state embedding to form DRL state embedding and fed into the DRL policy for navigation.
    }
    \label{fig:frame}

\vspace{-0.5cm}
\end{figure*}

\section{Methodology}

This article addresses the problem of vision-based robot navigation in crowded environments.
Our method (Fig.~\ref{fig:frame}) consists of three primary components: a feature extraction module, a feature fusion module, and a DRL network. 
Specifically, for the feature extraction module, the spatio-temporal encoder extracts occupancy features of the scene, while the I$^2$Former extracts intention-related features from the poses of surrounding pedestrians. These features are then concatenated to form intention-aware scene representations, which are further fused with robot state features to form the state embedding for the DRL network. Finally, the DRL network predicts navigation actions to enable collision-free navigation in dynamic environments.

\subsection{Problem Formulation}
The navigation problem extends the general visual navigation objective: learn to navigate along a collision-free path towards the goal by leveraging its egocentric visual observations, with the additional objective of maintaining an appropriate social distance from pedestrians.
In particular, this problem can be formulated as a partially observed Markov decision process. 
The partial observation $\mathbf{o}^t$ consists of egocentric images and robot states, and action $\mathbf{a}^t$ represents linear and angular velocity commanded to the robot. 
The navigation policy $\pi_{\theta}$ models the conditional distribution of actions given observations, denoted as $\pi_{\theta}(\mathbf{a}^t|\mathbf{o}^t)$. 
Our goal is to optimize the navigation policy $\pi_{\theta}$ with DRL, which can be achieved by maximizing the general objective $L(\theta)$:
\begin{align}
L(\theta)=\mathbb{E}_{\pi_{\theta}}\left[\sum_{t=0}^{\infty} \gamma^t r^t \right]
\end{align}
where $\gamma$ is the discount rate, and $r^t$ is the reward at time $t$ that evaluates the safety and efficiency of the navigation.

We summarize the commonly used notations in the following. 
Uppercase letters denote constants such as dimensions, while bold lowercase letters denote latent vectors in neural networks. 
For instance, BEV features are denoted by $\mathbf{x}$, and $\mathbf{z}$ denotes the state embedding used by the policy network. 
Functions such as neural networks are denoted by $\phi$, and their learnable parameters are denoted by $\mathbf{W}$.

\subsection{Intention-Aware Scene Representations}

In this subsection, we introduce the detailed structure for learning intention-aware scene representations, which consists of two main modules, the spatio-temporal encoder and the I$^2$Former, as shown in Fig.~\ref{fig:algorithm}. Our method takes the robot’s partial observation $\mathbf{o}^t$ as input, which comprises three components: the RGB-D data, the pedestrian poses detected from the RGB-D data, and robot states.

\subsubsection{Spatio-Temporal Encoder}

For the perception of the surroundings, we transform the multi-view visual observations into BEV feature maps as intermediate representations for downstream encoding, which provide a spatially consistent representation that facilitates reasoning about nearby pedestrians and obstacles in complex environments.
Fig~\ref{fig:algorithm} illustrates the pipeline for extracting spatio-temporal BEV features $\mathbf{s}^t_\text{bev}$ and encoding them into the embedding $\mathbf{z}^t_{\text{bev}}$ using a 2D convolutional BEV encoder.

% Our spatio-temporal encoder is inspired by the Fiery model\cite{hu2021fiery}.
% At timestep $t$, RGB-D observations from $N_c$ cameras,
% $[\mathbf{I}^t, \mathbf{D}^t] \in \mathbb{R}^{N_c \times (3+1) \times H \times W}$,
% are processed by a pre-trained ResNet-18 \cite{he2016deep} to extract RGB features $\mathbf{e}^t \in\mathbb{R}^{N_c \times C\times H_c\times W_c}$, while depth images are downsampled to a matching resolution $\mathbf{d}^t$.
% Using the camera intrinsics and extrinsics, along with the downsampled depth images $\mathbf{d}^t$, the RGB features $\mathbf{e}^t$ are lifted to 3D into a common reference frame (the inertial center of the ego-robot at time $t$). 
% The resulting 3D features are then sum-pooled along the vertical dimension to form the current BEV features at timestep $t$, $\mathbf{x}^t_\text{bev}\in\mathbb{R}^{C\times H_b\times W_b}$, with ($H_b$, $W_b$) = (120, 200) the spatial extent of the BEV feature.

Our spatio-temporal encoder is inspired by the Fiery model\cite{hu2021fiery}.
At timestep $t$, multi-view images from $N_c$ RGB-D cameras on the robot are collected and processed using a pre-trained ResNet-18 \cite{he2016deep}, producing the feature map $\mathbf{e}^t \in \mathbb{R}^{N_c \times C \times H_c \times W_c}$. 
The corresponding depth images are downsampled to the same resolution, yielding $\mathbf{d}^t \in \mathbb{R}^{N_c \times H_c \times W_c}$.
Then, we lift the image features into 3D using the measured depth under known camera parameters, and project them into a unified ego-centric coordinate frame at time $t$. 
The resulting 3D features are collapsed along the vertical dimension through sum pooling, producing the local BEV feature map 
$\mathbf{x}^t_{\text{bev}} \in \mathbb{R}^{C \times H_b \times W_b}$, 
where $H_b =120, W_b = 200$.

% we process BEV features across $T$ historical frames by first spatially aligning each to the current frame $t$ via the robot's egomotion 
Subsequently, to construct spatio-temporal BEV features, we aggregate BEV representations from a temporal window $\tau = \{t-T, \dots, t\}$. 
The historical BEV features are first aligned to the current frame at time $t$ by compensating for ego-motion using geometric transformations from the robot's trajectory.
These aligned feature maps, alongside the current feature map $\mathbf{x}^t_\text{bev}$, are then fed into the temporal encoder to extract spatio-temporal representations.
This module is denoted as $\mathcal{F}_\tau$ that performs temporal alignment and encoding:
\begin{align}
    \mathbf{s}^t_\text{bev}=\mathcal{F}_\tau({\mathbf{x}}^{t-T}_\text{bev},\dots,{\mathbf{x}}^{t-1}_\text{bev},\mathbf{x}^t_\text{bev})\in\mathbb{R}^{C\times H_b\times W_b}.
\end{align}
The temporal encoder is implemented as a 3D convolutional network and pre-trained on the nuScenes dataset \cite{caesar2020nuscenes}, which provides the multi-view camera setup required for BEV representation and contains pedestrian data.

Finally, the spatio-temporal BEV features are fed into the BEV encoder to extract scene representations $\mathbf{z}^{t}_\text{bev}$ : \begin{align}
    \mathbf{z}^{t}_\text{bev} = \phi_\text{bev}(\mathbf{s}^t_\text{bev};\mathbf{W}_{\text{bev}}),
\end{align}
where $\phi_\text{bev}$ is a 2D convolutional network with residual connection and 2D pooling layers.

\subsubsection{Intent-Interact Former}

\begin{figure}[!t]
\vspace{0.25cm}
\centering
    \includegraphics[width=1.00\linewidth]{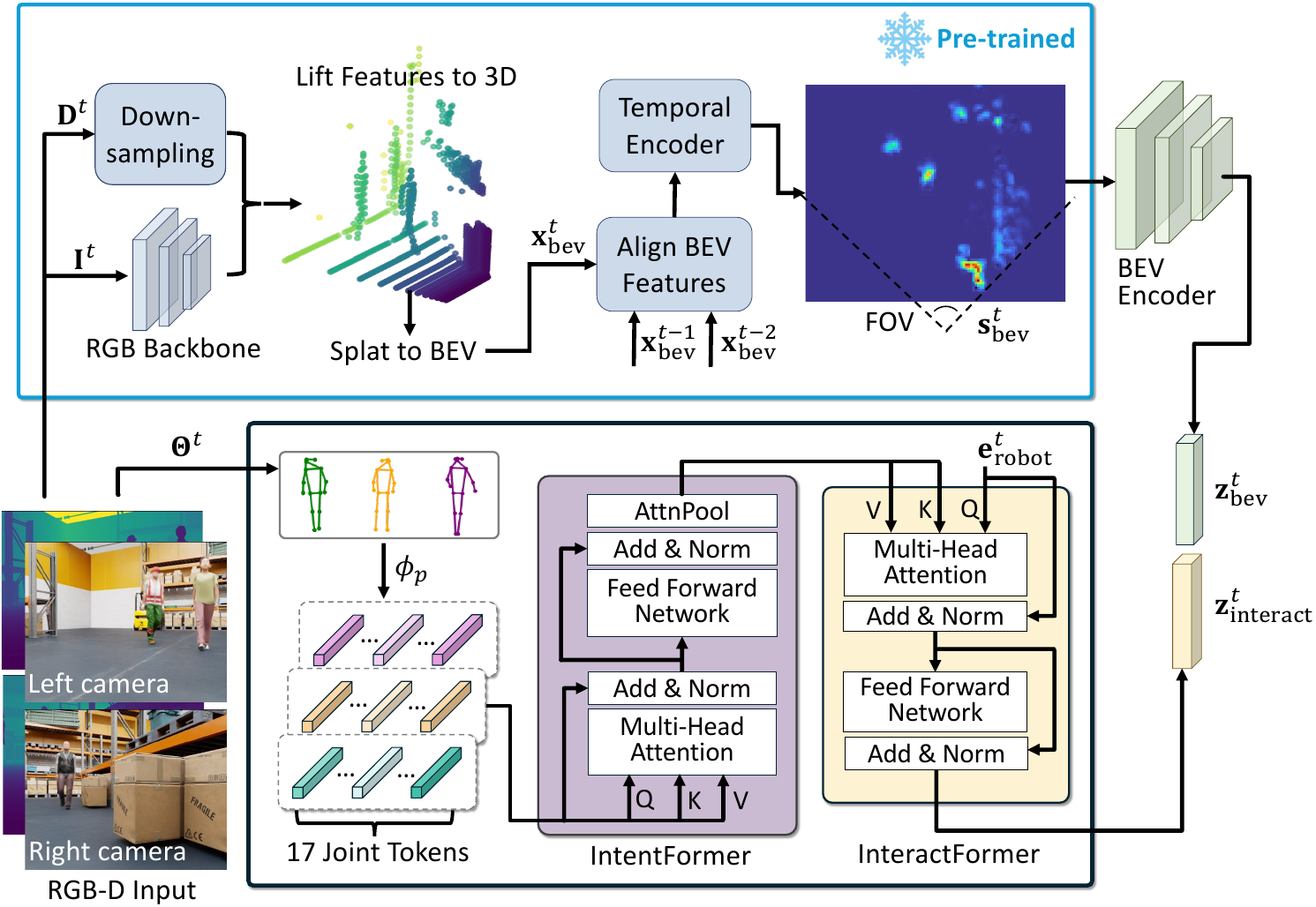}
    \caption{
    Our method includes two key components: a spatio-temporal encoder and the I$^2$Former. 
    The spatio-temporal encoder extracts scene features that implicitly capture the occupancy of both static and dynamic objects in the surrounding environment, while the I$^2$Former extracts intention-related features from the poses of surrounding pedestrians. 
    These features are then concatenated to form intention-aware scene representations.
    }
    \label{fig:algorithm}
    \vspace{-0.5cm}
\end{figure}

To capture behavioral intentions from human poses, we design I$^2$Former (Fig.~\ref{fig:algorithm}), which comprises four modules: a pose encoder, an $\text{I}^2$-states encoder, IntentFormer, and InteractFormer. 
We first detect 2D poses from RGB images using Ultralytics YOLO \cite{yolov8_ultralytics}, which achieves highly accurate pose estimation even under occlusions.
These 2D poses are then lifted to 3D and transformed into the robot’s coordinate frame using camera intrinsics, extrinsics, and depth images, yielding $\mathbf{\Theta}^t=[\mathbf{p}^t_1,...,\mathbf{p}^t_{N_p}]\in \mathbb{R}^{N_p\times 17 \times 3}$ for $N_p$ pedestrians with 17 joints.

To map the raw 3D coordinates into higher-dimensional embeddings, we apply an MLP pose encoder $\phi_p$ with ReLU activations:
\begin{equation}
\hat{\mathbf{\Theta}}^t =\phi_{p}(\mathbf{\Theta}^t;\mathbf{W}_{\phi_p})
\end{equation}
\rv{Crucially, occluded keypoints are zero-padded \cite{salzmann2023robots}. Through its global context modeling, the Transformer inherently directs its attention weights toward detected keypoints, maintaining robust representations even with incomplete poses.}

Human pose reflects a pedestrian’s next motion state. 
By modeling the relationships among all joints, we can infer behavioral intentions, such as moving forward, turning, or yielding. 
To this end, we design the IntentFormer, which leverages multi-head self-attention (MHSA) to capture these joint relationships, enabling the model to understand implicit behavioral intentions. 
It also incorporates a Feed-Forward Network (FFN), residual connections, and LayerNorm (LN). 
Additionally, attention pooling (AttnPool) is applied at the final stage, allowing the model to adaptively weigh the importance of different joints and improve its understanding of human motions and intentions. 
The IntentFormer outputs implicit intention features $\mathbf{f}^t_\text{ped}$, which can be formulated as:
\setlength{\jot}{5pt}
\begin{gather}
\hat{\mathbf{f}}^t_\text{ped} = \text{LN}(\text{MHSA}(\hat{\mathbf{\Theta}}^t) + \hat{\mathbf{\Theta}}^t), \\
\mathbf{f}^t_\text{ped} = \text{AttnPool}(\text{LN}(\text{FFN}(\hat{\mathbf{f}}^t_\text{ped}) + \hat{\mathbf{f}}^t_\text{ped})).
\end{gather}

After obtaining the intention features of each pedestrian, a module is needed to associate the robot with them to capture interactions between the robot and surrounding pedestrians. 
To this end, we design the InteractFormer based on multi-head cross-attention (MHCA), where the robot’s state embedding serves as the query to attend to the surrounding pedestrians’ features, enabling the robot to form a global understanding of their behavioral intentions. 
The InteractFormer outputs interact representations $\mathbf{z}^t_\text{interact}$, which can be expressed as:
\setlength{\jot}{5pt}
\begin{gather}
\hat{\mathbf{z}}^t_\text{interact} = \text{LN}(\text{MHCA}(\mathbf{e}^t_\text{robot}, \mathbf{f}^t_\text{ped})+\mathbf{e}^t_\text{robot}), \\
\mathbf{z}^t_\text{interact} = \text{LN}(\text{FFN}(\hat{\mathbf{z}}^t_\text{interact})+\hat{\mathbf{z}}^t_\text{interact}),
\end{gather}
where $\mathbf{e}^t_\text{robot} = \phi_{sp}(\mathbf{s}^t; \mathbf{W}_{\phi_{sp}})$ denotes the robot state embedding, and the $\text{I}^2$-states encoder $\phi_{sp}$ is implemented as an MLP with ReLU activations.

\subsubsection{Feature Fusion}

To obtain the DRL state embedding $\mathbf{z}^t$, we first encode the robot state $\mathbf{s}^t = [g^t_x, g^t_y, d^t_g, v^t_x, v^t_y]$ that comprises the goal direction $[g^t_x, g^t_y]$, distance to the goal $d^t_g$, and velocity $[v^t_x, v^t_y]$ into the state embedding. 
This encoded state is then concatenated with the intention-aware scene representations and fused:
\begin{gather}
    \mathbf{z}^t_\text{state} = \phi_s(\mathbf{s}^t; \mathbf{W}_{\phi_s})
    \\
    \mathbf{z}^t = \phi_f(\text{Concat}(\mathbf{z}^t_\text{bev}, \mathbf{z}^t_\text{interact}, \mathbf{z}^t_\text{state}); \mathbf{W}_{\phi_f}).
\end{gather}
Both the state encoder $\phi_s$ and the fusion network $\phi_f$ are implemented as MLPs with ReLU activations.

\subsection{Deep Reinforcement Learning}
We adopt the proximal policy optimization (PPO) \cite{schulman2017proximal} algorithm for online training of our policy. 
During training, all the modules $\phi(\cdot)$ with weights $\mathbf{W}$ are jointly optimized with the DRL policy network. 
Since the RGB backbone and the temporal encoder of the spatio-temporal encoder are pretrained on external datasets, they are kept frozen during DRL training. 
This pretraining strategy enhances the robustness of our scene representations and the training stability.

The reward function is designed to guide the robot toward safe, collision-free navigation in dynamic and complex environments. According to \cite{xie2023drl, xu2025navrl}, it should provide dense feedback at every step, along with clear terminal signals that indicate success or failure. Therefore, we design the following reward function, which encourages the robot to move toward the goal while proactively avoiding pedestrians:
\begin{align}
r^t_\text{nav} = \begin{cases}
20, & \mathrm{if}\ d^t_g \leq \rho_\text{robot} \\
-20, & \mathrm{else\ if}\ d^t_o\leq \rho_\text{robot} \\
0.5(d^t_o-0.9), & \mathrm{else\ if}\  \rho_\text{robot} < d^t_o < 0.9 \\
3.2(d^{t-1}_g-d^{t}_g), & \mathrm{otherwise},
\end{cases}
\end{align}
where $\rho_\text{robot}$ is the radius of the robot, $d^t_g$ is the distance between the robot and its goal at time $t$, and $d^t_o$ is the minimum distance between the robot and any pedstrian or obstacle at time $t$. Unlike \cite{samsani2021socially, xie2023drl}, we do not design a complex reward function based on obstacle-avoidance strategies or human–robot interaction patterns, which would require careful manual tuning. Those works explicitly craft such rewards so that the neural network can learn concepts like human private space through training; in contrast, our scene representations already implicitly encode human intentions, obviating the need for finely hand-tuned reward terms. Since policies optimized via DRL may exhibit jitter, we also include a trajectory-smoothing reward following \cite{xie2023drl, xu2025navrl}, defined as follows:
\begin{align}
r^t_{\omega} = \begin{cases}
-0.1|\omega^t_z|, & \mathrm{if}\ |\omega^t_z |> 1.0 \\
0, & \mathrm{otherwise},
\end{cases}
\end{align}
where $\omega^t_z$ denotes the robot’s angular velocity at time step $t$.  
Therefore, our overall reward is the sum of these two components:
$
    r^t=r^t_\text{nav}+r^t_{\omega}.
$
\vspace{-0.1cm}

\section{Simulation Experiments}

\subsection{Simulation Implementation}
\begin{figure}[!t]
\vspace{0.25cm}
\centering{
    \includegraphics[width=0.97\linewidth]{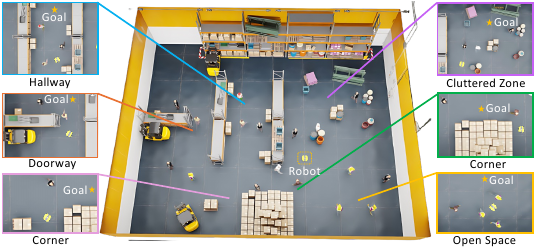}
    \label{fig:parrel_a}
}

\caption{Policy training environment in SocNav-Gym, featuring common social scenarios and providing diverse training data for DRL.}
\label{fig:socnav-gym}
\vspace{-0.5cm}
\end{figure}

During the simulation phase, we utilize a Clearpath Dingo robot with a maximum velocity of 1.0 m/s.
The robot is equipped with two Intel RealSense D435 RGB-D cameras, each with a depth range of [0.3, 10] m, providing a combined field of view of approximately 140°. 
\rv{Pedestrians move according to the Social Force Model (SFM) \cite{helbing1995social} toward fully randomized target destinations. Combined with the natural pedestrian animations rendered by Isaac Sim, these settings guarantee that our simulated crowd interactions closely mirror actual real-world scenarios.}
Fig~\ref{fig:socnav-gym} illustrates the training environment in SocNav-Gym, which covers various common social navigation scenarios such as hallways, corners, cluttered areas, and dense crowds in open spaces. 
In addition, SocNav-Gym also provides diverse testing environments, including hospital, office, and warehouse scenarios. In each episode, the robot's start and target positions are randomized to promote adaptation to diverse navigation scenarios.
% Based on this setting, we train two different DRL-based navigation policies: one with the full framework, and one without the I$^2$Former, thereby enabling a fair ablation study.

\subsection{Crowd Navigation}
\label{sec:crowdnav}

\begin{figure*}[!t]
\centering

\subfloat[Office lobby]{
    \includegraphics[width=0.31\linewidth]{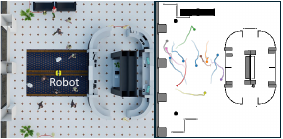}
    \label{fig:parrel_b}
}
\subfloat[Hospital corridor]{
    \includegraphics[width=0.30\linewidth]{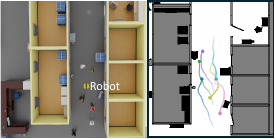}
    \label{fig:parrel_c}
}
\subfloat[Warehouse]{
    \includegraphics[width=0.31\linewidth]{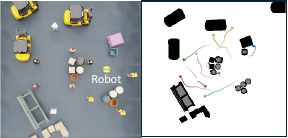}
    \label{fig:parrel_d}
}

\caption{
    Experimental environments for navigation policy testing. (a) Office lobby with a width of 7.0 m. (b) Hospital corridor with a width of 4.0 m. (c) Warehouse with a width of 2.5 m.
}
\label{fig:crowdnav-scene}
% \vspace{-0.5cm}
\end{figure*}

\begin{table*}[htbp]
\footnotesize
\centering
\vspace{-0.3cm}
\caption{Comparison Results in Simulated Crowd Navigation}
\label{table:crowd_navigation}
\setlength{\tabcolsep}{7pt}

\begin{tabular}{c|l|cccc|cccc|cccc}
\toprule
\textbf{Density} & \textbf{Method} &
\multicolumn{4}{c|}{\textbf{Office Lobby}} &
\multicolumn{4}{c|}{\textbf{Hospital Corridor}} &
\multicolumn{4}{c}{\textbf{Warehouse}} \\
& &
SR$\uparrow$ & NT$\downarrow$ & PL$\downarrow$ & TPZ$\downarrow$ &
SR$\uparrow$ & NT$\downarrow$ & PL$\downarrow$ & TPZ$\downarrow$ &
SR$\uparrow$ & NT$\downarrow$ & PL$\downarrow$ & TPZ$\downarrow$
\\
\midrule

% ================= LOW ====================
\multirow{7}{*}{\textbf{Low}}
& \textbf{Ours}
& \textbf{0.95} & 9.03 & \textbf{5.87} & \textbf{0.92}
& \textbf{0.88} & 9.82 & 6.19 & \textbf{1.98}
& \textbf{0.87} & 9.77 & \textbf{6.15} & \textbf{1.83}
\\
& \textbf{Ours (w/o I$^2$)}
& 0.93 & 9.22 & 5.99 & 1.79
& 0.84 & 9.30 & 6.32 & 2.69
& 0.79 & 10.77 & 6.67 & 2.22
\\
& \textbf{Ours (w/o BEV)}
& 0.92 & 9.15 & 5.95 & 1.57
& 0.81 & 9.97 & 6.97 & 2.54
& 0.74 & 10.85 & 6.71 & 2.12
\\
& DRL-VO
& 0.93 & \textbf{8.32} & 5.91 & 1.29
& 0.80 & \textbf{8.60} & \textbf{6.08} & 3.08
& 0.80 & \textbf{9.67} & 6.38 & 2.33
\\
& SARL*-OM
& 0.85 & 8.89 & 6.31 & 1.23
& 0.75 & 13.00 & 9.10 & 3.87
& 0.75 & 10.16 & 6.40 & 1.88
\\
& ViNT
& 0.81 & 10.55 & 6.11 & 2.01
& 0.69 & 11.13 & 6.21 & 3.15
& 0.69 & 11.25 & 6.65 & 3.72
\\
& DWA
& 0.92 & 12.48 & 6.24 & 1.66
& 0.87 & 11.34 & 6.01 & 2.57
& 0.77 & 13.16 & 6.71 & 3.42
\\
\midrule

% ================= HIGH ====================
\multirow{7}{*}{\textbf{High}}
& \textbf{Ours}
& \textbf{0.91} & 9.20 & 6.18 & \textbf{1.42}
& \textbf{0.84} & 9.83 & 6.31 & \textbf{2.55}
& \textbf{0.80} & 11.15 & 7.02 & \textbf{2.83}
\\
& \textbf{Ours (w/o I$^2$)}
& 0.85 & 9.29 & 5.89 & 2.04
& 0.80 & 9.75 & 6.34 & 3.04
& 0.75 & 11.08 & 6.74 & 3.58
\\
& \textbf{Ours (w/o BEV)}
& 0.83 & 9.31 & 6.14 & 1.65
& 0.77 & 9.91 & 6.67 & 2.89
& 0.73 & 11.45 & 6.97 & 3.39
\\
& DRL-VO
& 0.83 & \textbf{8.51} & \textbf{5.95} & 2.54
& 0.72 & \textbf{8.86} & \textbf{6.28} & 3.01
& 0.71 & \textbf{10.89} & \textbf{6.68} & 3.71
\\
& SARL*-OM
& 0.67 & 9.04 & 6.35 & 2.05
& 0.67 & 14.90 & 9.95 & 3.13
& 0.57 & 11.50 & 7.61 & 3.17
\\
& ViNT
& 0.71 & 11.05 & 6.01 & 2.91
& 0.65 & 13.51 & 6.79 & 4.01
& 0.51 & 12.51 & 7.12 & 5.02
\\
& DWA
& 0.75 & 12.94 & 6.86 & 2.88
& 0.75 & 12.46 & 6.35 & 3.50
& 0.72 & 15.18 & 7.60 & 4.79
\\

\bottomrule
\end{tabular}
% \vspace{-0.5cm}
\end{table*}

\begin{figure*}[!t]
\centering
    \includegraphics[width=0.95\linewidth]{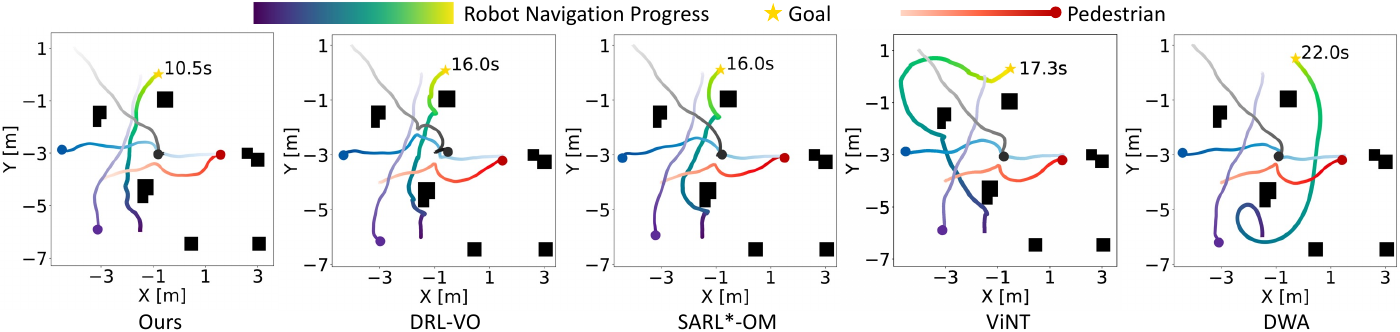}
    \caption{
    Example trajectories for the compared policies with nine static obstacles and four SFM agents. 
    The robot trajectory is color-coded with the viridis colormap to indicate the navigation timesteps. Black squares denote obstacles.
    }
    \label{fig:crowd-nav-traj}
    \vspace{-0.5cm}
\end{figure*}

We first conduct a series of crowd navigation experiments, where the initial distances between the robot and the goal are around 6.0 m.
According to benchmark \cite{stratton2024characterizing}, social navigation scenarios can be categorized based on crowd density and scene width. 
Therefore, we design three test scenarios with varying widths: an office lobby, a hospital corridor, and a warehouse with cluttered obstacles. 
Their widths are 7.0 m, 4.0 m, and 2.5 m, as illustrated in Fig.~\ref{fig:crowdnav-scene}. 
In addition, we set two crowd density levels: low (0.1) and high (0.2) pedestrians/m$^{2}$.
For comparison, we utilize four widely used metrics:
\begin{enumerate}
\item \textbf{Success rate (SR)}: the proportion of non-collision trials among all trials.%the fraction of collision-free trials.
\item \textbf{Navigation time (NT)}: the average time taken to reach the goal in successful trials.
\item \textbf{Path length (PL)}: the average length traveled in successful trials.
\item \textbf{Time in private zone (TPZ)}: the average time the robot spent in pedestrians’ private zones (distance $<$ 0.8 m) during successful trials.
\end{enumerate}

To evaluate the effects of both components, we perform an ablation study comparing the full method with two variants: one without the I$^2$Former and one replacing the spatio-temporal encoder with a CNN that encodes the occupancy map (OM).
For the comparative experiments, the following methods are used as baselines: DRL-VO \cite{xie2023drl}, SARL*-OM \cite{samsani2021socially,liu2020robot}, ViNT \cite{shah2023vint}, and DWA \cite{fox2002dynamic}. DRL-VO and SARL*-OM are DRL-based, targeting navigation efficiency and pedestrian comfort, respectively, with SARL*-OM combining the local OM \cite{liu2020robot} and danger-zone modeling \cite{samsani2021socially}.
ViNT is a visual navigation foundation model with obstacle avoidance capability.
DWA is a model-based collision avoidance controller.
For each method and configuration, we run three trials with 25 random goals in the corresponding test environment. Results are summarized in Table~\ref{table:crowd_navigation}.

Across environments and density levels, our method consistently outperforms the ablated variants, demonstrating stronger overall performance in crowded and constrained scenarios. Removing the I$^2$Former lowers SR and increases time in pedestrians’ private zones, reflecting impaired intention inference. Removing the BEV module reduces scene awareness, leading to less flexible navigation, frequent pauses, and higher intrusion into pedestrian space. These results show that BEV representations improve navigation by providing spatial and semantic features of the environment, while I$^2$Former enhances navigation by encoding pedestrians’ intentions.

When evaluated against the baselines, our method maintains the highest SR, with clear advantages in dense and constrained environments. It also navigates more efficiently, with shorter times and paths, while keeping the lowest TPZ, reflecting safer behavior. Even in narrower, more crowded settings, its performance degrades more slowly than baselines. Notably, the version without I$^2$Former, which relies only on the BEV visual representation, performs comparably to DRL-VO, which requires a full map, complete pedestrian states, and LiDAR fusion. Compared with ViNT, another visual navigation model, our approach surpasses it in all performance metrics.

For qualitative evaluation, we record robot trajectories with different navigation policies, as shown in Fig.~\ref{fig:crowd-nav-traj}. 
All policies enable the robot to reach the goal, but show notable differences. 
DWA’s conservative policy leads to detours and the longest navigation time. 
ViNT demonstrates inflexible navigation, resulting in excessively long paths and collisions with pedestrians.
SARL*-OM exhibits a rather rigid navigation strategy and often gets stuck or collides with static obstacles.
DRL-VO emphasizes efficiency but can disturb nearby pedestrians and hesitates near obstacles. 
In contrast, our method effectively perceives obstacles and initiates early avoidance maneuvers, while also respecting pedestrian social space, minimizing disturbance, and achieving the shortest time.

\subsection{Long-horizon Comparison}

\begin{figure*}[!t]
\centering
\subfloat[long-distance navigation in hospital]{
    \includegraphics[width=0.47\linewidth]{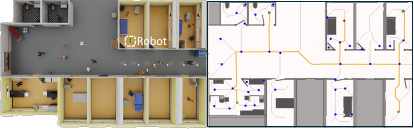}
    \label{fig:task-scene(a)}
}
\subfloat[long-distance navigation in office]{
    \includegraphics[width=0.51\linewidth]{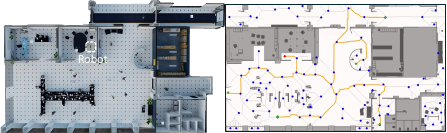}
    \label{fig:task-scene(b)}
}
\caption{
Two long-distance Navigation Scenarios. 
Each scenario uses a generalized voronoi graph to generate a topological map. 
In each map, \textcolor{green}{green dots} indicate possible goal points, \textcolor{red}{red dot} denotes start point, \textcolor{blue}{blue dots} represent waypoints, and \textcolor{orange}{orange segments} show the navigation paths. Each trajectory is defined as the robot starting from the start point and navigating along the paths to reach a randomly selected goal point.
}
\label{fig:task-scene}
\vspace{-0.5cm}
\end{figure*}

\begin{figure*}[!t]
\centering
\subfloat[Navigation in an outdoor gym]{
    \includegraphics[width=0.32\linewidth]{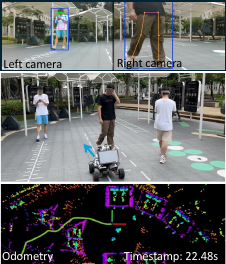}
    \label{fig:hardware-gym}
}
\subfloat[Navigation in a station]{
    \includegraphics[width=0.32\linewidth]{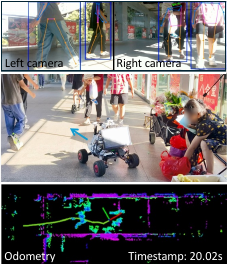}
    \label{fig:hardware-subway}
}
\subfloat[Navigation in a mall]{
    \includegraphics[width=0.32\linewidth]{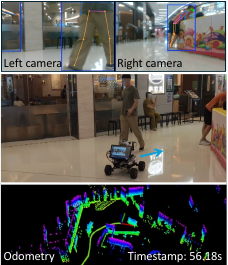}
    \label{fig:hardware-mall}
}
\caption{
    Snapshots and visualizations of the proposed method operating in three scenarios: an outdoor gym, a subway station, and a shopping mall.
    Each column shows the robot’s first view, the corresponding real-world scene, and odometry visualizations in RViz.
}
\label{fig:hardware}
\vspace{-0.5cm}
\end{figure*}

We then integrate the proposed algorithm with a topological map to evaluate crowd navigation within long horizon tasks. 
The topological map is generated by the generalized voronoi graph \cite{wang2020optimal}. 
We design two different scenarios, as illustrated in Fig.~\ref{fig:task-scene}. 
Both scenarios are large-scale, supporting navigation over 20 m. Compared with the previous crowd navigation settings, they better reflect real-world conditions, requiring the robot to navigate through corridors, doorways, and rooms commonly found in real world.
The evaluation metrics and baselines are consistent with those in Subsection~\ref{sec:crowdnav}, and each navigation task is repeated 25 times.

Across both environments, our full method outperforms the versions without the I$^2$Former and without the BEV module. Removing the I$^2$Former lowers SR and increases intrusions into pedestrians’ private zones, highlighting the value of capturing human motion intentions. Removing the BEV module reduces navigation flexibility and efficiency, while also affecting personal space compliance. In the office, our method achieves higher SR and fewer intrusions while remaining efficient. In the narrower hospital environment, its benefits are more evident, enabling reliable navigation.

When benchmarked against baselines, our method consistently achieves superior performance in SR, efficiency, and minimal intrusion. In the office environment, it reaches an SR of 0.95, \rv{outperforming the others}, while reducing TPZ from 7.96 (SARL*-OM) to 1.70, indicating safer and less intrusive navigation. In the hospital environment, it achieves an SR of 0.79 with the lowest TPZ (0.42), demonstrating robust and safe navigation even in narrower passages. Compared with ViNT, which is also a visual navigation method, our approach consistently achieves better results across all metrics.

\begin{table}[htbp]
\centering
\caption{Long-Horizon Comparison Results}
\label{table:long-distance navigation}
\setlength{\tabcolsep}{8pt}
\begin{tabular}{lccccc}
\toprule
\multirow{1}{*}{Env} & Method & SR$\uparrow$ & NT$\downarrow$ & PL$\downarrow$ & TPZ$\downarrow$ \\
\midrule
\multirow{7}{*}{\shortstack{Office}}
& \textbf{Ours}     & \textbf{0.95}   & \textbf{40.82} & 26.53   & \textbf{1.70} \\
& \textbf{Ours (w/o I$^2$)}     & 0.80   & 49.39 & \textbf{24.21}   & 2.10 \\
& \textbf{Ours (w/o BEV)} &0.83	&47.01	&28.21	&2.20\\
& DRL-VO      & 0.75   & 41.02 & 25.02   & 1.80 \\
& SARL*-OM & 0.33   & 68.64 & 39.81   & 7.96 \\
& ViNT &0.41	&55.87	&32.49	&5.61\\
& DWA        & 0.37   & 61.21 & 27.32   & 3.80 \\
\midrule
\multirow{7}{*}{\shortstack{Hospital}}
& \textbf{Ours}     & \textbf{0.79}   & 40.89 & 22.49   & \textbf{0.42} \\
& \textbf{Ours (w/o I$^2$)}      & 0.72   & 37.15 & \textbf{20.06}   & 1.37 \\
& \textbf{Ours (w/o BEV)} &0.73	&42.34	&23.29	&1.21\\
& DRL-VO     & 0.67   & \textbf{35.31} & 21.90  & 1.48 \\
& SARL*-OM     & 0.42   & 46.90 & 30.02   & 1.65 \\
& ViNT &0.56	&42.51	&24.38	&1.89\\
& DWA     & 0.50   & 45.19 & 25.25   & 2.15 \\
\bottomrule
\end{tabular}
\vspace{-0.3cm}
\end{table}

\section{Real-World Validation}

Besides simulated evaluations, we conduct real-world experiments in complex scenes—including a gym, a subway station, and a shopping mall—to validate the robustness and applicability of our method, with full demonstrations provided in the Multimedia. \rv{Our policy achieves an inference rate of 15 Hz on the onboard computer's RTX 2060 GPU.}

In the outdoor gym, where pedestrians behave non-cooperatively and frequently block the robot’s path, the robot actively adjusts its heading to yield space for passing, resulting in smooth and socially compliant avoidance behaviors, as shown in Fig.~\ref{fig:hardware-gym}.
In the subway station, the robot navigates through dense crowds and constrained spaces; it reliably percepts limited free space and promptly steers toward safer regions, successfully avoiding both moving pedestrians and static obstacles, as illustrated in Fig.~\ref{fig:hardware-subway}.
In the shopping mall, the robot completes a 109.49 m long-distance navigation in a constraint environment; even when a pedestrian suddenly emerges from a blind corner, it rapidly adjusts its trajectory to maintain safe separation and stable motion, as shown in Fig.~\ref{fig:hardware-mall}, achieving an average speed of 0.76 m/s.
Additional results under occlusions are provided in Appendix~A.

\section{Conclusion}
This article presents iCrowdNav, a visual navigation framework designed for robots operating in populated and dynamic environments. 
Unlike existing methods that rely on oversimplified scene representations, our approach learns the intention-aware scene representations directly from egocentric vision, allowing the robot to infer human intention and preserve more navigation-relevant visual cues, thereby enabling efficient and safe navigation in challenging crowds.
We validate our method in simulation and real-world settings, showing safer and more robust navigation in crowded scenarios, with successful deployment on physical robots.
However, severe occlusions and the limited field of view of egocentric cameras make intention inference difficult in ultra-dense scenes. Future work will explore richer multi-modal scene representations to improve intention reasoning and further enhance navigation robustness.

% \section*{Acknowledgments}
% This research includes calculations carried out on HPC resources supported in part by the National Science Foundation through major research instrumentation grant number 1625061 and by the US Army Research Laboratory under contract number W911NF-16-2-0189.

% \bibliographystyle{IEEEtran}
\bibliographystyle{ieeetrs}
\bibliography{v1/ref.bib}

\end{document}